\documentclass[conference]{IEEEtran}
\IEEEoverridecommandlockouts
\usepackage{cite}
\usepackage{amsmath,amssymb,amsfonts}
\usepackage{algorithmic}
\usepackage{graphicx}
\usepackage{textcomp}
\usepackage{xcolor}
\usepackage{hyperref}
\usepackage{subcaption}
\usepackage{url}
\def\BibTeX{{\rm B\kern-.05em{\sc i\kern-.025em b}\kern-.08em
    T\kern-.1667em\lower.7ex\hbox{E}\kern-.125emX}}
\begin{document}

\title{A Heterogeneous UAV UGV Methodology for Aerial Tracking of Mini Autonomous Cars
}

\author{\IEEEauthorblockN{Santiago Fernandez}
\IEEEauthorblockA{\textit{Robotics and AI Lab} \\
\textit{Technological University of Uruguay}\\
Rivera, Uruguay\\
santiago.fernandez@estudiantes.utec.edu.uy}
\and
\IEEEauthorblockN{Angel da Silva}
\IEEEauthorblockA{\textit{Robotics and AI Lab} \\
\textit{Technological University of Uruguay}\\
Rivera, Uruguay\\
angel.dasilva@estudiantes.utec.edu.uy}
\and
\IEEEauthorblockN{Hiago Sodre}
\IEEEauthorblockA{\textit{Robotics and AI Lab} \\
\textit{Technological University of Uruguay}\\
Rivera, Uruguay \\
hiago.sodre@utec.edu.uy}
\and
\IEEEauthorblockN{Pablo Moraes}
\IEEEauthorblockA{\textit{Robotics and AI Lab} \\
\textit{Technological University of Uruguay}\\
Rivera, Uruguay \\
pablo.moraes@utec.edu.uy}
\and
\IEEEauthorblockN{Bruna Guterres}
\IEEEauthorblockA{\textit{Robotics and AI Lab} \\
\textit{Technological University of Uruguay}\\
Rivera, Uruguay \\
bruna.devargas@utec.edu.uy}
\and
\IEEEauthorblockN{Ricardo B. Grando}
\IEEEauthorblockA{\textit{Robotics and AI Lab} \\
\textit{Technological University of Uruguay}\\
Rivera, Uruguay \\
ricardo.bedin@utec.edu.uy}
}

\maketitle

\begin{abstract}
Aerial tracking of autonomous ground vehicles can provide useful situational awareness in environments where fixed external perception infrastructure is unavailable. However, visual-only tracking from an unmanned aerial vehicle (UAV) is limited by camera field of view, viewpoint changes, occlusions, and temporary loss of target visibility. This paper presents a heterogeneous UAV--UGV methodology for aerial tracking of small-scale autonomous ground vehicles in simulation. The system is implemented in ROS~2 and Gazebo Sim using a Tello-like UAV equipped with a front-facing RGB camera and multiple LIMO-like unmanned ground vehicles (UGVs). Three approaches are evaluated: a training-free color-marker baseline, a YOLO-based visual baseline, and a proposed UAV--UGV fusion method. The proposed method uses the same visual detector as the YOLO baseline, but combines visual detections with UGV state information to maintain the target estimate during visual gaps. The methods are evaluated across four scenarios: single-car tracking, crossing trajectories, occlusion, and UAV-assisted awareness. Results over 600 simulation runs show that the proposed fusion method achieves nearly continuous track coverage across all scenarios while reducing localization error relative to the YOLO baseline. Compared with the visual-only YOLO approach, the fusion method reduces RMSE by 37.3\%, 64.2\%, 62.6\%, and 61.8\% in the four evaluated scenarios, respectively. The method also enables conflict-awareness lead times of 10.20~s in the crossing scenario and 9.07~s in the UAV-assisted awareness scenario. These results indicate that heterogeneous UAV--UGV fusion can improve aerial tracking continuity and robustness when visual observations are intermittent.
\end{abstract}


\section{Introduction}
\label{sec:introduction}

Heterogeneous robotic teams combine platforms with complementary sensing, mobility, and computational capabilities. In air--ground systems, unmanned aerial vehicles (UAVs) provide an elevated viewpoint and wide-area perception, while unmanned ground vehicles (UGVs) offer stable ground-level motion, longer operational persistence, and direct interaction with the environment. This complementarity has motivated increasing interest in UAV--UGV cooperation for inspection, mapping, monitoring, search-and-rescue, and collaborative autonomy \cite{zhou2026airgroundsurvey,miller2022airground}.

Aerial tracking of autonomous ground vehicles can be particularly useful in environments where fixed external perception infrastructure is unavailable. This aerial perspective allows the UAV to simultaneously monitor multiple vehicles, detect potentially hazardous situations, and provide complementary information for trajectory coordination. This capability can support fleet monitoring, collision prevention, large-area inspection, and assistance to ground vehicles whose perception is limited by obstacles, short-range sensors, or unfavorable environmental conditions.

Despite these advantages, purely visual tracking is limited by the camera field of view, viewpoint changes, partial occlusions, and target appearance variations. These limitations become more critical in multi-robot scenarios where ground vehicles may cross paths, become partially occluded, or enter shared areas where early conflict awareness is required.

Real-time object detection has become a standard component of robotic perception systems. The YOLO (You Only Look Once) family is widely adopted because it combines accurate object localization with real-time inference \cite{redmon2016yolo}. Recent implementations such as YOLOv8 provide practical training and deployment pipelines for robotic applications \cite{jocher2023yolov8}. However, continuous target awareness cannot be achieved through visual detection alone. Visual observations become unavailable when the tracked UGV leaves the camera field of view or becomes temporarily occluded. Under these conditions, a UAV relying only on visual perception loses the target despite the detector providing accurate observations whenever the vehicle is visible.

\begin{figure*}[t] 
    \centering 
    \includegraphics[width=\textwidth]{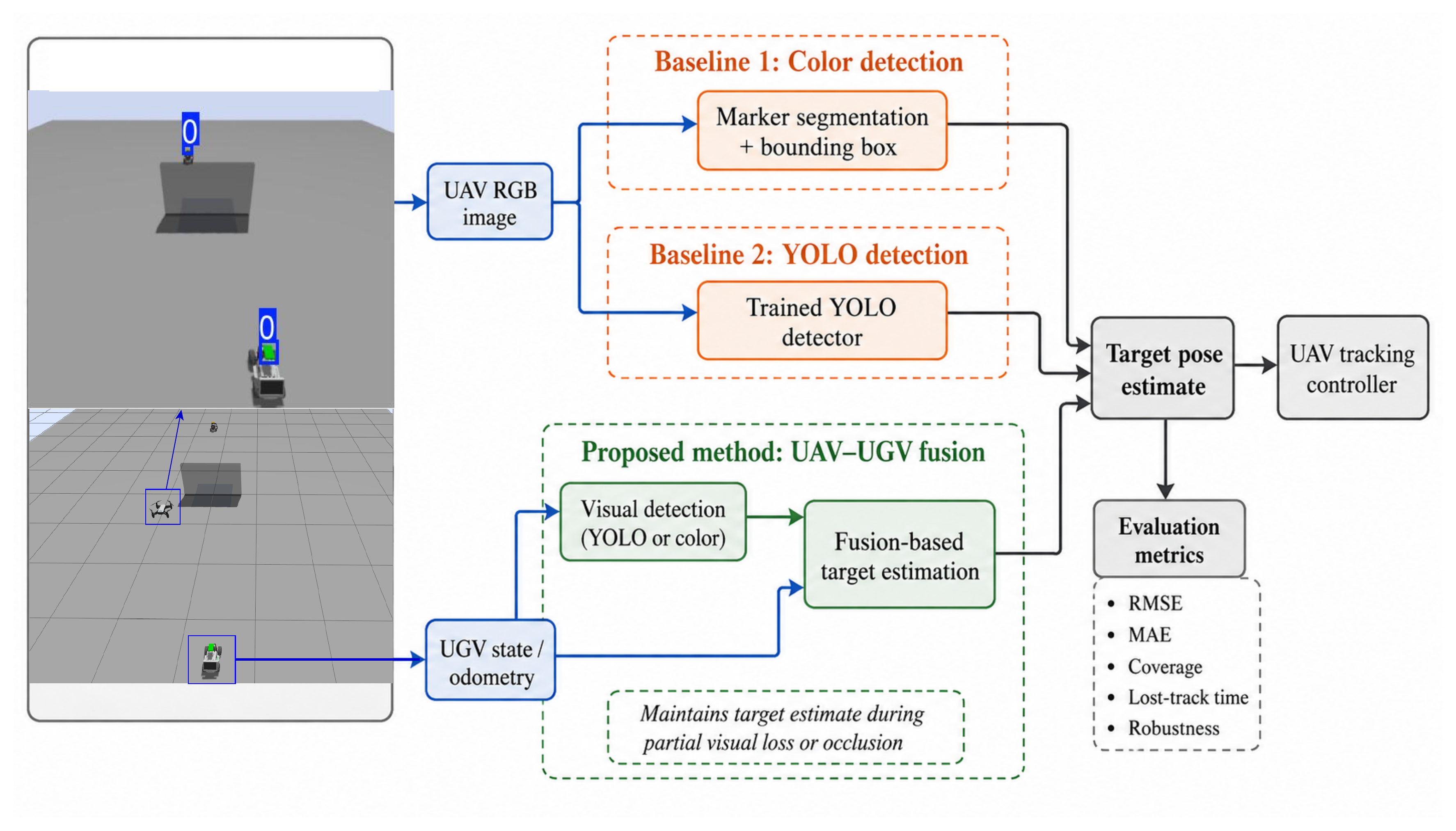} 
    \caption{Overview of the proposed heterogeneous UAV--UGV aerial tracking methodology. The UAV provides front RGB images for visual detection, while UGV state information is used by the proposed fusion method to maintain target estimates during visual gaps. The same simulation environment is used to evaluate the color baseline, the YOLO baseline, and the proposed UAV--UGV fusion approach. Code availability: \url{https://github.com/ricardoGrando/limo_tello_jazzy_spawn}. Video: \url{https://www.youtube.com/watch?v=f7xAwTxisKk}} \label{fig:methodology_framework} 
\end{figure*}

This paper proposes a heterogeneous UAV--UGV methodology for simulated aerial tracking of small-scale autonomous ground vehicles. The system employs a mini UAV with a front-facing RGB camera to observe UGVs in a ROS~2 and Gazebo Sim environment. Three tracking approaches are evaluated. The first is a training-free color baseline. The second is a visual baseline based on YOLO. The third is the proposed heterogeneous UAV--UGV approach that integrates visual perception with information shared by the ground vehicle. This comparison isolates the contribution of heterogeneous cooperation to aerial target tracking.Figure~\ref{fig:methodology_framework} summarizes the proposed heterogeneous UAV--UGV tracking methodology. The simulated environment provides two complementary information streams. The first is the UAV front RGB image. The second is the UGV state information. The RGB stream is processed by both visual baselines. The proposed method combines visual detections with UGV state information to estimate the target position when visual observations become unavailable. This design isolates the effect of heterogeneous information on tracking continuity.

The main contributions of this work are:
\begin{itemize}
    \item A ROS~2/Gazebo Sim methodology for heterogeneous UAV--UGV aerial tracking of small-scale autonomous ground vehicles.
    \item A comparison between a color-based visual baseline, a YOLO visual baseline, and a UAV--UGV fusion approach.
    \item An evaluation across four scenarios: single-car tracking, crossing trajectories, occlusion, and UAV-assisted awareness.
    \item Experimental evidence that the proposed fusion method increases track coverage to nearly 100\% while reducing localization error across all evaluated scenarios.
\end{itemize}
\section{Related Work}
\label{sec:related_work}

\subsection{UAV Target Tracking}

UAV target tracking has been studied as both an active and passive perception problem. In active tracking, the UAV adjusts its motion to maintain the target within its field of view, while passive tracking focuses on estimating information about the target from the available observations. State-estimation and fusion strategies, such as Kalman filtering, particle filtering, and learning-based tracking, are commonly used to improve robustness under noisy or intermittent observations \cite{wu2025uavsurvey}.

Vision-based UAV tracking systems often combine object detection with temporal filtering. Barisic et al. presented a vision-based system for real-time detection and following of UAVs using YOLO and a Kalman filter, showing that combining visual detections with dynamic estimation improves tracking stability \cite{barisic2022uavfollowing}. These works support the idea that visual detections are useful but should be complemented with temporal state estimation.

\subsection{YOLO-Based Robotic Perception}

YOLO is a family of single-stage detectors capable of estimating object locations and classes within a single inference step \cite{redmon2016yolo}. Its balance between detection accuracy and processing speed makes it particularly suitable for robotic applications requiring real-time visual perception. YOLOv8 is distributed through the Ultralytics framework, which provides pretrained models, training utilities, and deployment tools \cite{jocher2023yolov8}. Recent work has analyzed YOLOv8 and reported its practical advantages for real-time object detection, including its anchor-free design and efficient training pipeline \cite{yaseen2024yolov8}.

In UAV applications, YOLO-based detectors have been used for flying-object detection, small-object recognition, and lightweight onboard perception. Reis et al. evaluated YOLOv8 for real-time flying-object detection and emphasized challenges related to small target size, motion, occlusion, and cluttered backgrounds \cite{reis2023flyingobjects}. These challenges are also present in UAV--UGV tracking, where temporary occlusions or field-of-view limitations may interrupt visual observations of the target.

\subsection{Heterogeneous UAV--UGV Cooperation}

Heterogeneous UAV--UGV systems have been studied for cooperative localization, mapping, planning, and semantic collaboration. Recent surveys describe air--ground heterogeneous systems as a promising robotic paradigm because UAVs and UGVs provide complementary capabilities in perception, mobility, and mission execution \cite{zhou2026airgroundsurvey}. Other works have explored air--ground cooperation for semantic mapping and navigation, where aerial perception provides information that improves ground-robot decision-making \cite{miller2022airground}.

However, it remains important to evaluate under controlled conditions to what extent shared UGV state information can preserve aerial tracking continuity during temporary visual loss.

The present work follows this research direction but focuses specifically on aerial tracking of small-scale autonomous ground vehicles. Unlike a purely visual method, the proposed approach uses the same visual detector as the YOLO baseline and combines visual detections with UGV state information, including position, orientation, and velocity, to maintain continuous tracking during temporary visual loss. This design isolates the contribution of heterogeneous fusion from the contribution of the learned detector.

\section{Methodology}
\label{sec:methodology}

The proposed methodology (Figure \ref{fig:methodology_framework}) is implemented in ROS~2 Jazzy and Gazebo Sim. The simulated heterogeneous team consists of a mini UAV and one or more UGVs. The UAV is equipped with a front-facing RGB camera and observes the UGVs from an elevated viewpoint. The UGVs follow predefined trajectories according to the selected scenario, while the UAV provides aerial visual information. The simulation considers a Tello-like UAV and an Agilex-like UGV.

The tracking pipeline contains four main modules:
\begin{enumerate}
    \item \textbf{Scenario controller:} generates the motion patterns for each experimental scenario.
    \item \textbf{Visual detector:} processes the UAV RGB stream and outputs target detections.
    \item \textbf{Tracking module:} converts detections into estimated target states and maintains tracks.
    \item \textbf{Experiment logger and evaluator:} records samples and computes coverage, localization error, and conflict-awareness metrics.
\end{enumerate}

Three tracking methods are evaluated. These include a color baseline, a YOLO visual baseline, and the proposed UAV--UGV fusion approach, as described in the following subsections.

\subsection{Color Baseline}

The color baseline is a training-free visual method. It detects a colored marker placed on the top of the LIMO using RGB thresholding and connected-component filtering. This method is included as a classical baseline because it does not require a learned detector or a training dataset. However, it is expected to be sensitive to viewpoint changes, marker visibility, lighting, and geometric projection errors.

\subsection{YOLO Visual Baseline}

The YOLO baseline uses a trained YOLOv8 detector to identify the LIMO marker from the UAV RGB image. The model is trained using images collected from the simulation environment. To avoid inaccurate labels caused by simple geometric projection from the front camera, the final dataset labels are generated from the visible marker in the RGB image.

During evaluation, YOLO detections are converted into estimated target positions using a bounding-box-based front-camera range model. This method represents a learned visual baseline: it can localize the target accurately when the marker is visible, but it does not use UGV state information when detections are absent.

\subsection{Proposed UAV--UGV Fusion Method}

The proposed method uses the same trained YOLO detector as the YOLO baseline. Therefore, both methods share the same visual perception capability. The difference is that the proposed approach fuses YOLO detections with UGV state information.

When YOLO detections are available, the tracker updates the estimated target state using the visual measurement. When the UAV does not detect the target, the tracker maintains the target estimate using UGV state prediction. This mechanism allows the method to preserve track continuity during visual gaps caused by limited field of view, occlusion, or target motion outside the camera view.

This design isolates the contribution of fusion. Any improvement over the YOLO baseline is not due to a stronger detector, but to the use of heterogeneous UAV--UGV information.

\subsection{Experimental Scenarios}

Four scenarios are evaluated, as summarized in Table~\ref{tab:scenarios}. Each method is evaluated over 200 runs, composed of four scenarios, five random seeds, and ten trials per seed.

\begin{table}[t]
\centering
\caption{Evaluated scenarios.}
\label{tab:scenarios}
\begin{tabular}{lll}
\hline
\textbf{Scenario} & \textbf{Description} & \textbf{Main challenge} \\
\hline
S1 & Single car & Basic aerial tracking \\
S2 & Crossing & Interaction and crossing trajectories \\
S3 & Occlusion & Visual interruption and continuity \\
S4 & UAV-assisted awareness & Shared-zone awareness \\
\hline
\end{tabular}
\end{table}

\begin{figure*}[t]
    \centering
    \begin{subfigure}[t]{0.32\textwidth}
        \centering
        \includegraphics[width=\textwidth]{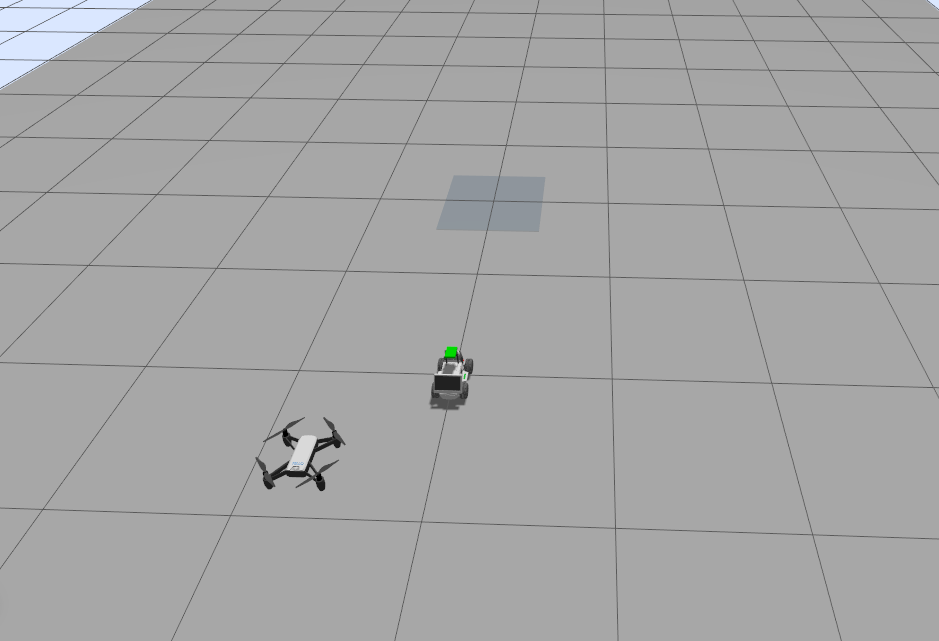}
        \caption{S1: single-car tracking.}
        \label{fig:scenario_s1}
    \end{subfigure}
    \hfill
    \begin{subfigure}[t]{0.32\textwidth}
        \centering
        \includegraphics[width=\textwidth]{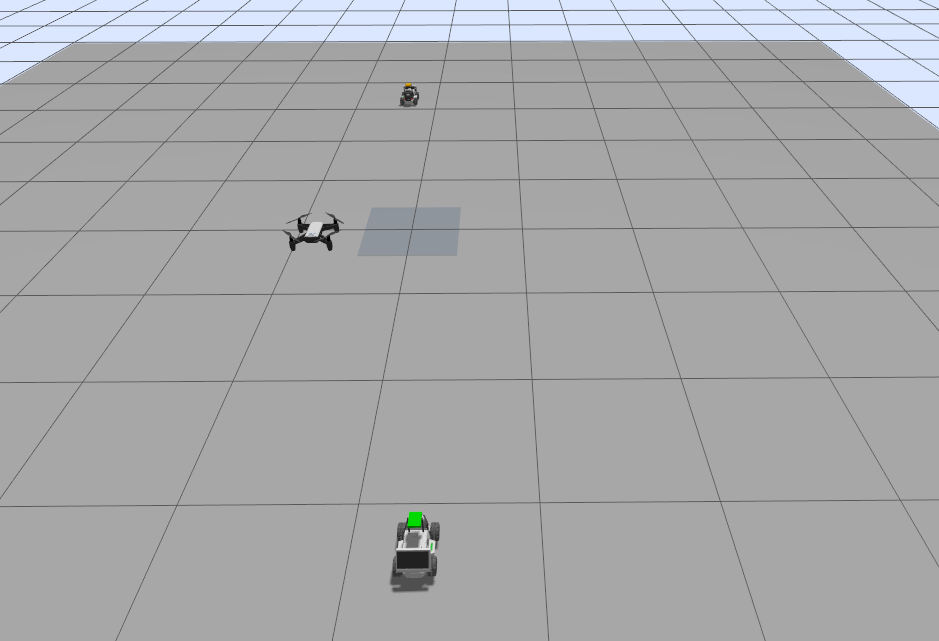}
        \caption{S2/S4: crossing and shared-zone awareness.}
        \label{fig:scenario_s2_s4}
    \end{subfigure}
    \hfill
    \begin{subfigure}[t]{0.32\textwidth}
        \centering
        \includegraphics[width=\textwidth]{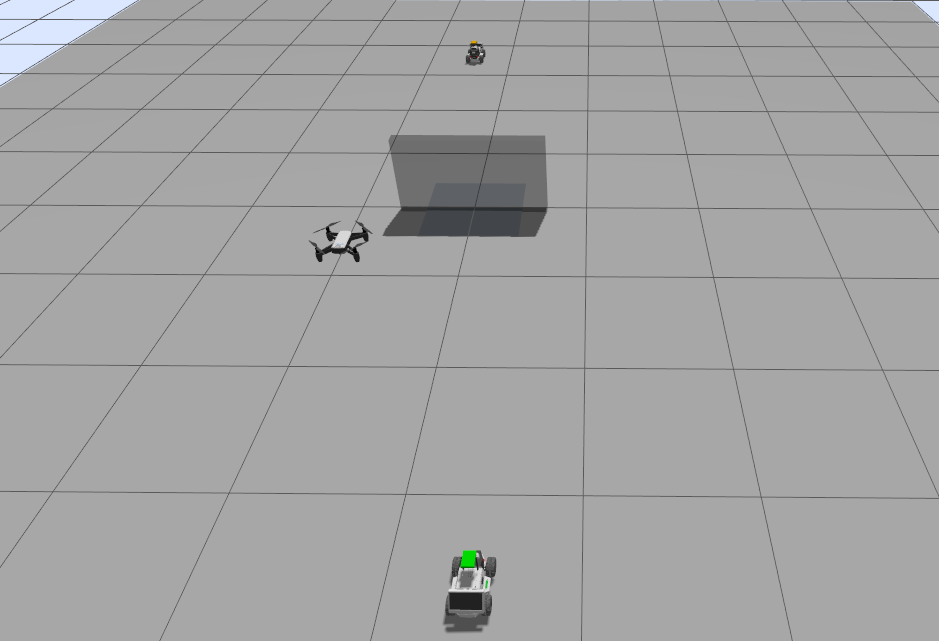}
        \caption{S3: occlusion scenario.}
        \label{fig:scenario_s3}
    \end{subfigure}
    \caption{Simulated evaluation scenarios in Gazebo. The Tello-like UAV observes one or more AgileX LIMO UGVs from an aerial viewpoint. The scenarios evaluate single-target tracking, multi-vehicle interaction, occlusion, and UAV-assisted shared-zone awareness.}
    \label{fig:experimental_scenarios}
\end{figure*}

Figure~\ref{fig:experimental_scenarios} illustrates the simulated environments used to evaluate the proposed heterogeneous UAV--UGV tracking methodology. The scenarios were designed to progressively increase the difficulty of the aerial tracking task. In S1, a single LIMO vehicle is tracked by the Tello-like UAV in an open environment. In S2, two ground vehicles move through a shared area, producing a crossing interaction that requires the tracking system to preserve target continuity. Scenario S3 introduces an occluding obstacle between the UAV and the ground vehicles, creating temporary visual loss. Finally, S4 uses the same multi-vehicle shared-zone setting to evaluate UAV-assisted awareness, where continuous tracking is required to anticipate possible interactions between vehicles.

\subsection{Metrics}

The tracking methods were evaluated using the following metrics:
\begin{itemize}
    \item \textbf{Detection coverage:} fraction of samples with a visual detection.
    \item \textbf{Track coverage:} fraction of samples with an active target track.
    \item \textbf{RMSE:} root mean squared position error.
    \item \textbf{MAE:} mean absolute position error.
    \item \textbf{Position standard deviation:} variability of the position error.
    \item \textbf{False positives:} number of incorrect detections.
    \item \textbf{ID switches:} number of identity assignment changes.
    \item \textbf{Conflict lead time:} time between conflict awareness and the potential interaction event.
\end{itemize}

Detection coverage measures the availability of visual observations. Track coverage measures whether the system maintains an active target estimate over time. The proposed fusion method is intended to improve track continuity rather than increase the number of visual detections. Therefore, both metrics are required to evaluate its performance.

\section{Results}
\label{sec:results}

\subsection{Evaluation Validity}

All three methods completed the full evaluation matrix. Each method produced 200 valid runs, with 50 runs per scenario. The YOLO baseline and the proposed fusion method loaded the trained YOLO model correctly in all runs, with no fallback to the color proxy detector. No false positives or ID switches were observed. Table~\ref{tab:validity} summarizes the evaluation validity.

\begin{table}[t]
\centering
\caption{Evaluation validity checks.}
\label{tab:validity}
\begin{tabular}{lccc}
\hline
\textbf{Check} & \textbf{Color} & \textbf{YOLO} & \textbf{Fusion} \\
\hline
Total runs & 200 & 200 & 200 \\
Runs per scenario & 50 & 50 & 50 \\
Metric files generated & 200 & 200 & 200 \\
YOLO loaded correctly & N/A & 200 & 200 \\
Ultralytics failures & N/A & 0 & 0 \\
Color proxy fallback & N/A & 0 & 0 \\
False positives & 0 & 0 & 0 \\
ID switches & 0 & 0 & 0 \\
\hline
\end{tabular}
\end{table}

\subsection{Quantitative Results}

Table~\ref{tab:main_results} presents the full quantitative comparison. The color baseline provides a simple visual reference, but its localization error is high. The YOLO baseline substantially improves localization accuracy when the target is visible. The proposed UAV-UGV fusion method achieves the best overall performance, especially in track coverage and localization error.

\begin{table*}[t]
\centering
\caption{Main quantitative results across all scenarios. Values are reported as mean $\pm$ standard deviation over 50 runs per scenario.}
\label{tab:main_results}
\resizebox{\textwidth}{!}{
\begin{tabular}{llcccc}
\hline
\textbf{Scenario} & \textbf{Method} & \textbf{Detection coverage} & \textbf{Track coverage} & \textbf{RMSE (m)} & \textbf{MAE (m)} \\
\hline
S1 single car & Color baseline & $\mathbf{68.3 \pm 1.3\%}$ & $68.2 \pm 1.3\%$ & $3.640 \pm 0.002$ & $3.640 \pm 0.002$ \\
S1 single car & YOLO baseline & $64.9 \pm 13.8\%$ & $63.1 \pm 16.2\%$ & $0.022 \pm 0.017$ & $0.017 \pm 0.017$ \\
S1 single car & UAV--UGV fusion & $63.0 \pm 16.3\%$ & $\mathbf{99.9 \pm 0.2\%}$ & $\mathbf{0.014 \pm 0.006}$ & $\mathbf{0.010 \pm 0.005}$ \\
\hline
S2 crossing & Color baseline & $\mathbf{58.0 \pm 5.9\%}$ & $59.9 \pm 6.0\%$ & $6.143 \pm 0.189$ & $5.405 \pm 0.156$ \\
S2 crossing & YOLO baseline & $34.9 \pm 0.5\%$ & $35.3 \pm 0.5\%$ & $2.531 \pm 0.023$ & $2.294 \pm 0.023$ \\
S2 crossing & UAV--UGV fusion & $31.5 \pm 8.4\%$ & $\mathbf{99.9 \pm 0.1\%}$ & $\mathbf{0.905 \pm 0.232}$ & $\mathbf{0.476 \pm 0.126}$ \\
\hline
S3 occlusion & Color baseline & $26.7 \pm 26.9\%$ & $27.8 \pm 28.1\%$ & $6.436 \pm 0.008$ & $5.786 \pm 0.009$ \\
S3 occlusion & YOLO baseline & $\mathbf{35.6 \pm 0.1\%}$ & $38.2 \pm 0.2\%$ & $2.600 \pm 0.007$ & $2.379 \pm 0.010$ \\
S3 occlusion & UAV--UGV fusion & $32.3 \pm 6.9\%$ & $\mathbf{99.9 \pm 0.1\%}$ & $\mathbf{0.974 \pm 0.204}$ & $\mathbf{0.542 \pm 0.116}$ \\
\hline
S4 UAV-assisted awareness & Color baseline & $0.0 \pm 0.0\%$ & $0.0 \pm 0.0\%$ & N/A & N/A \\
S4 UAV-assisted awareness & YOLO baseline & $\mathbf{34.8 \pm 0.3\%}$ & $34.8 \pm 0.3\%$ & $1.987 \pm 0.010$ & $1.952 \pm 0.011$ \\
S4 UAV-assisted awareness & UAV--UGV fusion & $34.8 \pm 0.4\%$ & $\mathbf{99.9 \pm 0.1\%}$ & $\mathbf{0.760 \pm 0.006}$ & $\mathbf{0.443 \pm 0.006}$ \\
\hline
\end{tabular}
}
\end{table*}

\begin{figure}[t] 
    \centering 
    \includegraphics[width=\columnwidth]{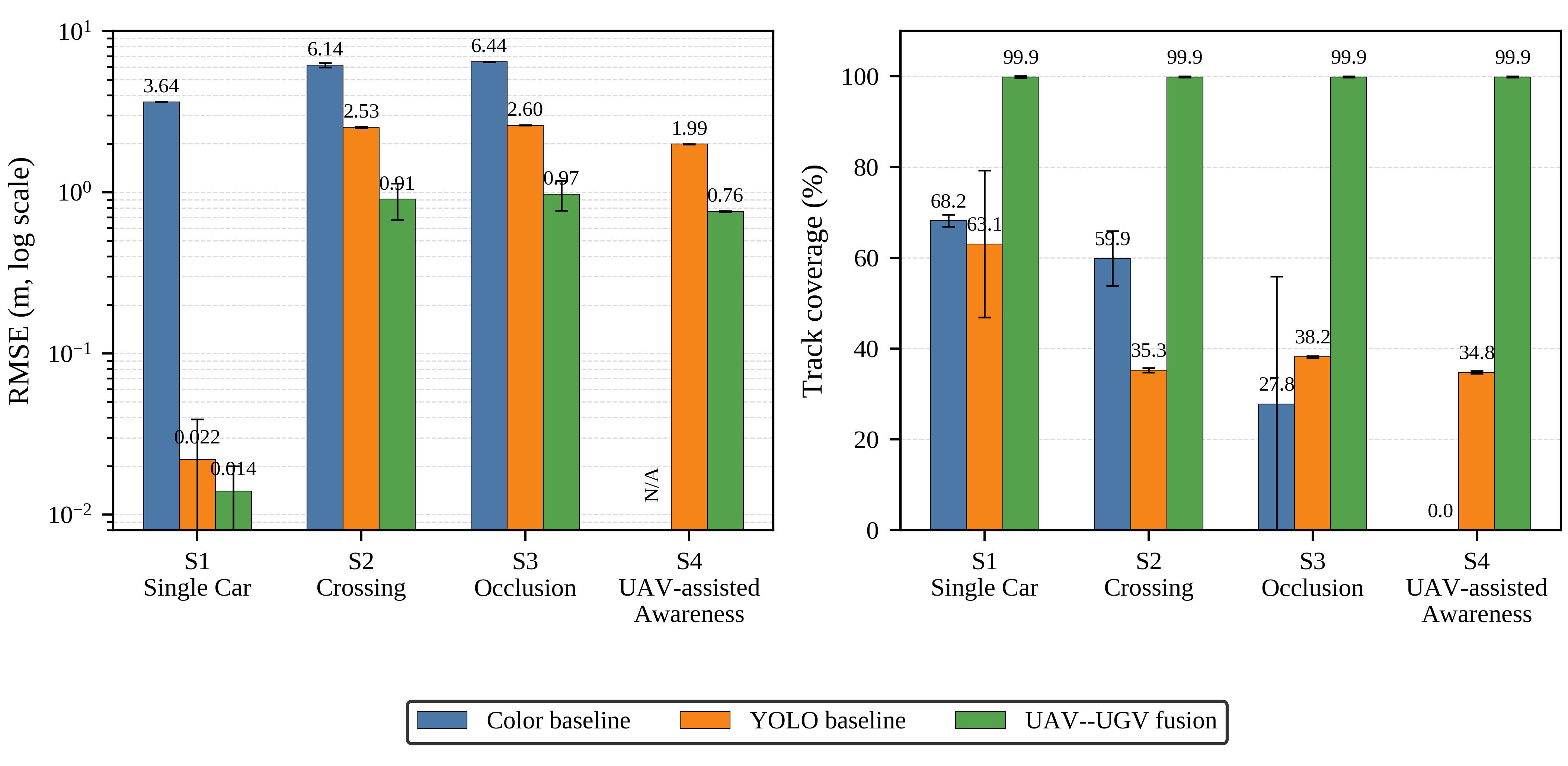} 
    \caption{Quantitative comparison across scenarios using the collected evaluation metrics. The proposed UAV--UGV fusion method reduces localization error and increases track coverage compared with the color and YOLO visual baselines.} 
    \label{fig:quantitative_comparison} 
\end{figure}

Figure~\ref{fig:quantitative_comparison} summarizes the main quantitative trends obtained from the full evaluation matrix. The left plot compares localization error, while the right plot compares track coverage across the four scenarios. The results are consistent with Table~\ref{tab:main_results}: the color baseline is able to detect the marker in some scenarios, but its localization error remains high; the YOLO baseline substantially improves localization accuracy when the target is visible; and the proposed UAV--UGV fusion method achieves the best overall balance between accuracy and continuity. In particular, the fusion method obtains the lowest RMSE and the highest track coverage in all scenarios, indicating that combining visual detections with UGV state information improves the robustness of aerial tracking.

\begin{figure}[t] 
    \centering 
    \includegraphics[width=0.9\columnwidth]{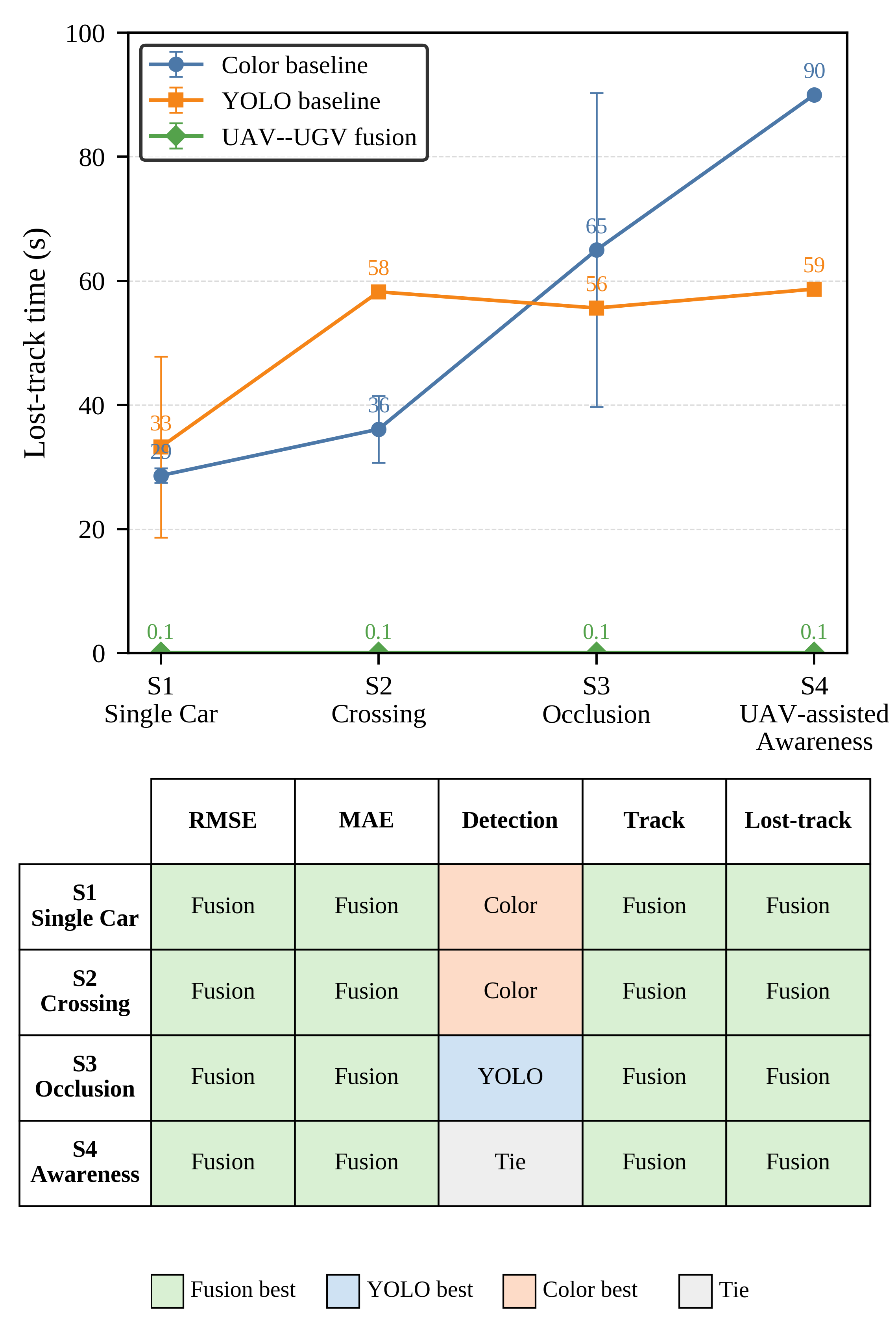} 
    \caption{Temporal robustness comparison based on lost-track time estimated from the collected track coverage. The proposed UAV--UGV fusion method maintains lower lost-track time across all scenarios.} 
    \label{fig:robustness_temporal} 
\end{figure}

Figure~\ref{fig:robustness_temporal} analyzes the same behavior from a temporal perspective by estimating the lost-track time from the measured track coverage. The visual-only baselines lose the target for longer periods as scenario complexity increases, especially in the crossing, occlusion, and shared-awareness scenarios. In contrast, the proposed fusion method maintains a substantially lower lost-track time because UGV state prediction preserves the target estimate during visual gaps. This result supports the main hypothesis of the paper: visual detection alone is useful when the target is visible, but heterogeneous UAV--UGV fusion is necessary to maintain continuous target awareness under intermittent observations.

\subsection{Improvement Over the YOLO Baseline}

Table~\ref{tab:improvements} reports the relative improvement of the proposed method over the YOLO visual baseline. Since both methods use the same trained YOLO detector, these gains isolate the effect of adding UGV state information to the tracking pipeline.

\begin{table}[t]
\centering
\caption{Improvement of the proposed method over the YOLO baseline.}
\label{tab:improvements}
\begin{tabular}{lccc}
\hline
\textbf{Scenario} & \textbf{Track gain} & \textbf{RMSE reduction} & \textbf{MAE reduction} \\
\hline
S1 & +36.7 pp & 37.3\% & 39.3\% \\
S2 & +64.5 pp & 64.2\% & 79.2\% \\
S3 & +61.7 pp & 62.6\% & 77.2\% \\
S4 & +65.1 pp & 61.8\% & 77.3\% \\
\hline
\end{tabular}
\end{table}

The largest improvement appears in track coverage. In S2, S3, and S4, the YOLO baseline tracks the target for only about one third of the trial because visual detections are available only during part of the trajectory. The proposed fusion method maintains nearly continuous tracking by using UGV state prediction whenever visual detections are unavailable. This also reduces localization error, particularly in the more challenging multi-vehicle and occlusion scenarios.

\subsection{Conflict Awareness}

The interaction scenarios S2 and S4 also evaluate conflict-awareness lead time. The visual baselines did not provide reliable conflict-awareness estimates, while the proposed fusion method produced early warnings in both scenarios. Table~\ref{tab:conflict} summarizes these results.

\begin{table}[t]
\centering
\caption{Conflict-awareness lead time in interaction scenarios.}
\label{tab:conflict}
\begin{tabular}{lccc}
\hline
\textbf{Scenario} & \textbf{Color} & \textbf{YOLO} & \textbf{Fusion} \\
\hline
S2 crossing & N/A & N/A & $\mathbf{10.20 \pm 0.22}$ s \\
S4 awareness & N/A & N/A & $\mathbf{9.07 \pm 0.06}$ s \\
\hline
\end{tabular}
\end{table}

This result shows that the proposed method improves not only target localization, but also situational awareness in scenarios where interaction between ground vehicles matters. By maintaining a continuous target estimate, the fusion approach can anticipate potential interactions earlier than visual-only baselines.

\subsection{Discussion}

The color baseline provides a useful training-free reference, but its position estimates are inaccurate under perspective changes and multi-robot configurations. Although it reaches moderate detection coverage in S1 and S2, its RMSE remains much higher than the learned and fusion-based approaches. In S4, the color baseline fails completely, indicating that simple marker segmentation is not sufficiently robust for the shared-awareness setting.

The YOLO baseline substantially improves localization accuracy when the target is visible. In S1, it achieves an RMSE of approximately $0.022$~m, showing that the learned detector and calibrated bounding-box range model can localize the marker accurately under favorable visibility. However, the YOLO baseline remains limited by the UAV camera field of view. In S2, S3, and S4, track coverage remains close to 35--38\%, reflecting the portions of the trajectory in which the target is actually observed by the UAV.

The proposed UAV--UGV fusion method overcomes this limitation by using UGV state information to maintain the target estimate during visual gaps. As a result, it achieves nearly 100\% track coverage across all scenarios and reduces RMSE relative to the YOLO baseline in every case. The largest gains occur in the crossing, occlusion, and shared-awareness scenarios, where visual-only tracking is most affected by temporary loss of visibility.
\section{Conclusion}
\label{sec:conclusion}

This paper presented a heterogeneous UAV--UGV methodology for aerial tracking of mini autonomous cars. The system was implemented in ROS~2 and Gazebo Sim using a Tello-like UAV and multiple UGVs. Three tracking approaches were compared. These included a training-free color baseline, a YOLO visual baseline, and a UAV--UGV fusion method.

The experiments showed that visual detection alone is sufficient while the target remains visible. Track continuity decreased when the target left the camera field of view or became occluded. The proposed fusion method maintained target estimates during these visual gaps by combining visual detections with UGV state information. This strategy improved tracking continuity without modifying the visual detector.

The proposed method achieved higher track coverage than both visual baselines across all evaluated scenarios. The largest improvements were observed during crossing trajectories, occlusions, and shared-awareness scenarios. These results indicate that heterogeneous state sharing can improve aerial target awareness when visual observations are intermittent. The proposed methodology also provides a reproducible framework for evaluating heterogeneous UAV--UGV cooperation under controlled simulation conditions.

Future work will evaluate the proposed methodology on real robotic platforms. The influence of sensor noise and communication delays will also be investigated. The simulated UGV state will be replaced with onboard localization estimates. Additional experiments will address multi-target identity management. Future work will also investigate UAV motion planning strategies that actively preserve target visibility.

\bibliographystyle{./bibliography/IEEEtran}
\bibliography{./bibliography/IEEEabrv,./bibliography/main}

\end{document}